\documentclass[conference]{IEEEtran}

\usepackage{graphicx}
\usepackage{epstopdf}
\usepackage{amsthm,amsmath,amssymb}
\usepackage{mathrsfs}
\usepackage{setspace}
\usepackage{amsmath}
\usepackage{booktabs}
\usepackage{enumerate}
\usepackage{cite}
\usepackage{stfloats}
\usepackage{subfigure}

\usepackage{algorithm}
\usepackage{algorithmic}
\usepackage{multirow}
\usepackage{flafter}
\usepackage{bm}

\usepackage{url}
\usepackage{xcolor}
\usepackage{subfigure}
\usepackage{fancyhdr}
\usepackage{mdwmath}
\usepackage{mdwtab}
\usepackage{caption}
\usepackage{amsthm}
\usepackage{setspace}
\usepackage{tabularx}
\usepackage{tabu}
\usepackage{lipsum}

\usepackage[
top    = 2.71cm,
bottom = 1.05in,
left   = 0.6 in,
right  = 0.6 in]{geometry}

\newtheorem{remark}{Remark}
\allowdisplaybreaks[4]

\ifCLASSINFOpdf
\else
\fi

\begin{document}

\title{Trajectory and Passive Beamforming Design for IRS-aided Multi-Robot NOMA Indoor Networks}

\author{\IEEEauthorblockN{ Xinyu~Gao\IEEEauthorrefmark{1}, Yuanwei~Liu\IEEEauthorrefmark{1}, and Xidong~Mu\IEEEauthorrefmark{2}}  
\IEEEauthorblockA{\IEEEauthorrefmark{1} Queen Mary University of London, London, UK\\
\IEEEauthorrefmark{2} Beijing University of Posts and Telecommunications, Beijing, China\\
 }}

\maketitle

\begin{abstract}
  A novel intelligent reflecting surface (IRS)-aided multi-robot network is proposed, where multiple mobile wheeled robots are served by an access point (AP) through non-orthogonal multiple access (NOMA). The goal is to maximize the sum-rate of all robots by jointly optimizing trajectories and NOMA decoding orders of robots, reflecting coefficients of the IRS, and the power allocation of the AP, subject to the quality of service (QoS) of each robot. To tackle this problem, a dueling double deep Q-network (D$^{3}$QN) based algorithm is invoked for jointly determining the phase shift matrix and robots' trajectories. Specifically, the trajectories for robots contain a set of local optimal positions, which reveals that robots make the optimal decision at each step. Numerical results demonstrated that the proposed D$^{3}$QN algorithm outperforms the conventional algorithm, while the performance of IRS-NOMA network is better than the orthogonal multiple access (OMA) network.
  \end{abstract}

\IEEEpeerreviewmaketitle

\section{Introduction}
A robot is an intelligent machine that can work semi-autonomously or fully autonomously, which can bestead or replace humans in accomplishing dangerous, arduous, and complex tasks, furthermore, expand the scope of human activities and capabilities. Among them, an integration technology of cellular networks with robots, namely, a communication-aware connected robot technology, turns into an appealing heated topic for channeling mission completion in recent years. This technology can support reducing the complexity of local calculations when robots are adopted for repeatedly handling high dimensional data. And in the fifth-generation (5G) mobile networks era, data transmission rate, latency, and compatibility of large-scale device connections have all been ameliorated. Integrating 5G or beyond 5G cellular networks with connected robots\cite{IEEEhowto:Galambos} are expected to create breakthrough developments. Additionally, communication-aware connected multi-robot systems provide system redundancy and enhanced capability compared to the single robot systems, which improves mission execution efficiency.
\par
Despite the aforementioned benefits of connected robots, there are some challenges during the applications. On the one hand, with respect to the multi-robot networks, the conventional scheme is to allocate a single wireless resource to a robot, such as by frequency or time, however, which cannot guarantee the spectrum efficiency and stability of multi-user connection. As an amelioration to the conventional scheme, non-orthogonal multiple access (NOMA) \cite{IEEEhowto:Liu} technologies enhance spectrum utilization efficiency and increase system throughput. The core idea of NOMA is to opportunistically explore the users' different channel conditions to superimpose the signals of robots, thereby improving spectrum utilization efficiency. On the other hand, when the robots are in the designated positions, the transmission link between the AP and the robots is blocked by obstacles and compels interruption of signal transmission. Intelligent reflecting surfaces (IRSs)\cite{IEEEhowto:Wu} can be a potential candidate solution for addressing this problem and enhancing communication quality. IRS has the capability of proactively modifying the wireless communication links by controlling a large number of passive reflective elements, which are recognized as a promising technique to enhance both spectrum efficiency and energy efficiency of wireless networks. Thus, the employment of the IRS can provide a solution for addressing signal blockage.
\par
Recent years, IRS-aided networks have witnessed a significant improvement on spectrum efficiency and energy efficiency. The authors in \cite{IEEEhowto:Cui} developed a K-means-based online user clustering algorithm to reduce the computational complexity and derive the optimal power allocation policy in a closed form. In \cite{IEEEhowto:FFang}, an energy-efficient algorithm is proposed to yield a good tradeoff between the sum-rate maximization and total power consumption minimization, by maximizing the system energy efficiency by jointly optimizing the transmit beamforming at the BS and the reflecting beamforming at the IRS. In \cite{IEEEhowto:Xmu}, efficient algorithms are proposed to maximize the sum-rate of all users by jointly optimizing the active beamforming at the BS and the passive beamforming at the IRS, subject to successive interference cancellation decoding rate conditions and IRS reflecting elements constraints. The authors of \cite{IEEEhowto:CHuang} investigated the energy efficiency maximization problem in an IRS-assisted multiple-user multiple-input single-output (MISO) system. Also, To further improve the spectrum efficiency, NOMA technology were considered in the IRS-assisted communication. An semidefinite relaxation based solution in \cite{IEEEhowto:MZeng} is proposed to address maximizing the sum-rate of all users in an IRS-assisted uplink NOMA system.  The authors in \cite{IEEEhowto:Ni} proposed a novel framework of resource allocation in multi-cell IRS-aided NOMA networks, which is capable of being enhanced with the aid of the IRS, and the proper location of the IRS can also guarantee the trade-off between spectrum and energy efficiency. The effectiveness of IRS in NOMA system with respect to transmit power consumption is examined in \cite{IEEEhowto:HWANG}, which can significantly reduce the required transmit power. The authors of \cite{IEEEhowto:Hou} analyzed various system performances in an IRS-aided NOMA network, and provided useful design insights.
\par
Sparked by the above advantages of NOMA and IRS, in this paper, we explore the potential performance gain of the IRS-aided multi-robot network. Particularly, we propose a novel framework for multi-robot networks, where NOMA is employed at the AP for serving multiple robots, and an IRS is invoked to enhance communication efficiency and overcome the signal blockage. Based on this framework, a sum-rate maximization problem is formulated by jointly optimizing trajectories and NOMA decoding orders of robots, reflecting coefficients of the IRS, and the power allocation at the AP, subject to the quality of service (QoS) of each robot. Then, an efficient algorithm is developed jointly determining the phase shift matrix and robots' trajectories. Numerical results show that: 1) The proposed IRS-aided NOMA networks achieve significant gain compared to IRS-OMA and without-IRS-assisted schemes; and 2) The proposed D$^{3}$DN algorithm outperforms the conventional algorithm.

\section{System Model and Problem Formulation}

\begin{figure}[htbp] 
  \centering
  \includegraphics[height=1.8in,width=3in]{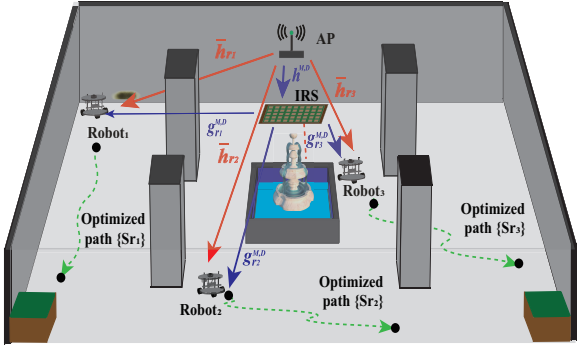}
  \caption{Illustration of the IRS aided multi-robots cruise system for the indoor environment.}
  \label{Illustration of the IRS aided multi robots cruise system for the indoor environment}
\end{figure}

As shown in Fig.~\ref{Illustration of the IRS aided multi robots cruise system for the indoor environment}, we focus our attention on a downlink IRS-aided multi-robot NOMA networks, which consists of one single-antenna AP which serves \emph{N} (an example of \emph{N = 3} is shown in Fig.~\ref{Illustration of the IRS aided multi robots cruise system for the indoor environment}) single-antenna mobile wheeled robots with the aid of an IRS with \emph{K} passive reflecting elements. We assume that the two-dimensional (2D) motion space for robots moving and rotating is approximately smooth without undulation, where the height \emph{$h_{r}$} of the robot (the height of the antenna) is treated as constant. For simplicity, the value involved in the precision of the motion space is ideal. Accordingly, the position of the AP is denoted as (\emph{$x_{A}$},\emph{$y_{A}$},\emph{$h_{A}$}). Note that for guaranteeing fairness, the IRS is located in the center of the ceiling in the environment. The passive reflecting elements \emph{K} in the IRS can be partitioned into \emph{M} sub-surfaces, while each sub-surfaces consists of \emph{$\widetilde{K}=K_{h}K_{v}$} elements. With the explored 2D motion space and the pre-defined three-dimension (3D) Cartesian coordinate system, the position of the IRS can be denoted as (\emph{$x_{I}$},\emph{$y_{I}$},\emph{$h_{I}$}). Additionally, the position of the robot \emph{$i$} is denoted as \emph{$S_{i}$} = (\emph{$x_{i}$},\emph{$y_{i}$},\emph{$h_{r}$}).
\par
In view of the deployment of the IRS, the composite received signal is the combination of two components, where the signals derived from AP-robot direct link, and the signals obtained from the AP-IRS-robot reflecting link. We denote baseband equivalent channels from the AP to the robot \emph{$i$}, the AP to the IRS and the IRS to the robot \emph{$i$} as \emph{$\overline{\pmb{h}}_{i}, i=\{1,2,\cdots,N\}$}, \emph{$(\pmb{h})^{H} \in \mathbb{C}^{1 \times K}$}, and \emph{$\pmb{g}_{i} \in \mathbb{C}^{K \times 1}, i=\{1,2,\cdots,N\}$}, respectively. Additionally, the distance-dependent channel path loss is modeled as \emph{$L_{u} = Cd_{u}^{-\gamma}, u=\{Ai,Ii,AI\}$} \cite{IEEEhowto:QWu}, where \emph{C} and \emph{$\gamma$} denote the path loss when the distance away from the AP is 1m, and the path loss factor, respectively. Denote the \emph{$q(t)$} and \emph{$\overline{q}=(\emph{$x_{I}$},\emph{$y_{I}$},\emph{$h_{I}$})$} as the positions of the robot and IRS. Thus, the three individual channels at location \emph{$q(t)$} for robot \emph{$i$} can be expressed as
\begin{align}\label{1}
  \overline{\pmb{h}}_{i}(q_{i}(t)) = L_{Ai}(q_{i}(t)) &\{\sqrt{\frac{1}{\alpha_{Ai}(q_{i}(t))+1}}\cdot \nonumber \\ & [\sqrt{\alpha_{Ai}(q_{i}(t))}\tilde{\pmb{l}}_{i}^{{\rm LoS}}(q_{i}(t))+\hat{\pmb{l}}_{i}^{{\rm NLoS}}]\},
\end{align}
\begin{align}\label{2}
  \pmb{g}_{i}(q_{i}(t)) = L_{Ii}(q_{i}(t)) &\{\sqrt{\frac{1}{\alpha_{Ii}(q_{i}(t))+1}}\cdot \nonumber \\ & [\sqrt{\alpha_{Ii}(q(t))}\tilde{\pmb{g}}_{i}^{{\rm LoS}}(q_{i}(t))+\hat{\pmb{g}}_{i}^{{\rm NLoS}}]\},
\end{align}
\begin{align}\label{3}
  \pmb{h} = L_{AI} \{\sqrt{\frac{1}{\alpha_{AI}+1}}[\sqrt{\alpha_{AI}}\tilde{\pmb{h}}^{{\rm LoS}}+\hat{\pmb{h}}^{{\rm NLoS}}]\},
\end{align}
\par
\noindent
where \emph{$\alpha_{u}, u=\{Ai,Ii,AI\}$}, \emph{$v^{{\rm LoS}}, v=\{\tilde{\pmb{l}}_{i},\tilde{\pmb{g}}_{i},\tilde{\pmb{h}}\}$}, \emph{$w^{{\rm NLoS}}, w=\{\hat{\pmb{l}}_{i},\hat{\pmb{g}}_{i},\hat{\pmb{h}}\}$} denote the Rician factor, deterministic line-of-sight (LoS) component and random non-line-of-sight (NLoS) Rayleigh fading components, respectively. Denote \emph{$\pmb{\Phi}(t) = {\rm diag}(\phi_{1}(t),\phi_{2}(t),\cdots,\phi_{K}(t))$} as the reflection coefficients matrix of the IRS, where \emph{$\phi_{k}(t) = \beta_{k}(t)e^{j\theta_{k}(t)}, k=\{1,2,3,\cdots,K\}$}. Additionally, the \emph{$|\beta_{k}(t)| = 1$} and \emph{$\theta_{k}(t) \in [0,2\pi)$} denote the amplitude and phase of \emph{k}-th element in the IRS. Thus, the effective channel from the AP to the robot \emph{$i$} is given by
\begin{align}\label{4}
  \pmb{H}_{i}(q_{i}(t)) = (\pmb{h})^{H} \pmb{\Phi}(t) \pmb{g}_{i}(q_{i}(t)) &+ \overline{\pmb{h}}_{i}(q_{i}(t)), \nonumber \\ &i=\{1,2,\cdots,N\}.
\end{align}
\par
The interference among robots cannot be negligible while one AP serves \emph{N} robots simultaneously. Thus, according to the fairness principle, the interference information needs to be employed for the received signal at robot \emph{$i$}. We consider the NOMA strategy to mitigate interference among robots. For the NOMA scheme, in addition to the superposition coding (SC) method, the successive interference cancelation (SIC) method should be leveraged for sharing the same time/frequency resources to all the robots.
\par
According to the NOMA principle, the SC method is applied at the AP. Let \emph{$S_{i}=\sqrt{p_{i}}s_{i}$} denote the transmitted signal for the robot \emph{$i,i=\{1,2,\cdots,N\}$}, while \emph{$s_{i},i=\{1,2,\cdots,N$\}} represents the transmitted information symbol for the robot \emph{$i,i=\{1,2,\cdots,N\}$}. It is worth noting that \emph{$S_{i}$} is satisfied \emph{$\mathbb{E}[|S_{i}|^{2}] = p_{i} \leq P_{i},i=\{1,2,\cdots,N\}$}, with \emph{$p_{i}$} and \emph{$P_{i}$} denoting the transmitted power and its maximum value of the robot \emph{$i,i=\{1,2,\cdots,N\}$}, respectively. SIC is applied for each robot to remove the interference. The robots with stronger channel power gain decode signals of other robots with weaker channel power gain priorily over decoding its own signal. Denote \emph{$O(i)$} as the decoding order of the robot \emph{$i$}. For any two robots \emph{$i$} and \emph{$j$}, \emph{$i \neq j, \hspace{0.25em} i,j=\{1,2,\cdots,N\}$}, if the decoding order satisfying \emph{$O(i) > O(j)$}, the recieved signal of robot \emph{$i$} in equation can be modeled as
\begin{align}\label{5}
  Y_{i}(q_{i}(t)) = \pmb{H}_{i}(q_{i}(t))S_{i} + \sum_{O(j) > O(i)} &\pmb{H}_{j}(q_{j}(t))S_{j} + n, \nonumber \\ &i=\{1,2,\cdots,N\},
\end{align}
\par
\noindent
where the \emph{$n \sim \mathcal{CN}(0,\sigma^{2})$} denotes the additive white Gaussian noise (AWGN) with average power \emph{$\sigma^{2}$}. For each robot \emph{$i,i=\{1,2,\cdots,N\}$}, the achievable rate can be denoted as \emph{$R_{i}$}. Then we denote \emph{$(\pmb{h})^{H} \pmb{\Phi} \pmb{g}_{i} = (\pmb{\upsilon})^{H}\pmb{\psi}_{i}$}, where \emph{$\pmb{\psi}_{i}$} = diag\{\emph{$(\pmb{h})^{H}$\}$\pmb{g}_{i}$}, \emph{$\pmb{\upsilon} = [\upsilon_{1}, \upsilon_{2}, \cdots, \upsilon_{K}]^{H}$}, and \emph{$\upsilon_{k} = e^{j\theta_{k}}$}. so the signal-to-interference-plus-noise ratio (SINR) of robot \emph{$i$} is given by
\begin{align}\label{6}
  \tau_{i}(q_{i}(t)) = \frac{[|(\pmb{\upsilon})^{H}\pmb{\psi}_{i}+\overline{\pmb{h}}_{i}|^{2}p_{i}]_{q_{i}(t)}}{\sum\limits_{O(j) > O(i)}[|(\pmb{\upsilon})^{H}\pmb{\psi}_{j}+\overline{\pmb{h}}_{j}|^{2}p_{j]_{q_{j}(t)}}+\sigma^{2}},
\end{align}
\par
\noindent
where the \emph{$\sigma^{2}$} denotes the variance of the AWGN. Then, according to the formula R=${\rm log}_{2}$(1+SINR), the achievable communication rate at robot \emph{$i$} can be expressed as
\begin{align}\label{7}
  R_{i}(q_{i}(t)) &= \nonumber \\
  &{\rm log}_{2}(1+\frac{[|(\pmb{\upsilon})^{H}\pmb{\psi}_{i}+\overline{\pmb{h}}_{i}|^{2}p_{i}]_{q_{i}(t)}}{\sum\limits_{O(j) > O(i)}[|(\pmb{\upsilon})^{H}\pmb{\psi}_{j}+\overline{\pmb{h}}_{j}|^{2}p_{j]_{q_{j}(t)}}+\sigma^{2}}).
\end{align}
\par
Let the \emph{$\mathcal{P}$} denote the total transmit power at the AP, where \emph{$\sum_{i=1}^{N}p_{i} \leq \mathcal{P}$}. According to the decoding order, the transmit power at robots \emph{$i$} and \emph{$j$} should satisfy the condition \emph{$p_{i} \geq p_{j}, O(j) < O(i)$}. 
\par
Our goal is that all the robots can achieve maximum sum-rate by jointly optimizing trajectories for robots, reflecting coefficients matrix, the decoding order, and the power allocation at the AP, subject to the quality of service (QoS) for all the robots. Thus, the optimization problem is formulated as
\begin{align}
  \max_{\pmb{\upsilon},\Omega,\{p_{i}\},\pmb{Q}} \hspace*{1em}& \sum\limits_{i=1}^{N} R_{i}(q_{i}(t)) \label{8}\\
  {\rm s.t.} \hspace*{1em}& R_{i}(q_{i}(t)) \geq \overline{R}, \tag{\ref{8}{a}} \label{8a}\\
  & |\pmb{\upsilon}| = 1, \tag{\ref{8}{b}} \label{8b}\\
  & [|(\pmb{\upsilon})^{H}\pmb{\psi}_{i}+\overline{\pmb{h}}_{i}|^{2}]_{q_{i}(t)}> [|(\pmb{\upsilon})^{H} \nonumber\\
  & [\pmb{\psi}_{j}+\overline{\pmb{h}}_{j}|^{2}]_{q_{j}(t)}, \hspace{0.5em} {\rm if} \hspace{0.5em} O_{i}> O_{j},\tag{\ref{8}{c}} \label{8c}\\
  & \Omega \in \pmb{\Pi}, \tag{\ref{8}{d}} \label{8d}\\
  & \sum_{i=1}^{N}p_{i} \leq \mathcal{P}, \tag{\ref{8}{e}} \label{8e} \\
  & q_{i}(t) \in \pmb{Q}_{i}, \tag{\ref{8}{f}} \label{8f}
\end{align}
\par
\noindent
where the \emph{$\overline{R}$}, \emph{$\pmb{\Pi}$} and \emph{$\pmb{Q} = [\pmb{Q}_{1},\pmb{Q}_{2},\cdots,\pmb{Q}_{N}]$} denote the minimal required communication rate for all the robots in the 2D explored space, the set of all the possible decoding orders and the set of trajectories for all the robots, respectively. Constraint (\ref{8a}) and constraint (\ref{8b}) are the QoS requirements for robot \emph{$i$} and the restraint for the IRS reflection coefficients. Constraint (\ref{8c}) and constraint (\ref{8d}) are the decoding conditions for the NOMA scheme. However, the main difficulty to solve the problem \eqref{8} involves the integer constraints for decoding order design owing to the following reasons. Firstly, according to the equations \eqref{1} - \eqref{4}, the channel model is position-dependent, which relies on no concave trajectory \emph{$\pmb{Q}_{i}$} and phase shift \emph{$\pmb{\upsilon}$}. Secondly, with respect to the continuous-time \emph{t}, infinite variables optimization is difficult to handle. Thirdly, the communication rate for each robot \emph{i} is generally not a continuous function in virtue of the position-dependent channel model. Fourthly, the individual channel consists of LoS and NLoS components, which is difficult to determine which components are included in the channel model established according to robot positions. Thus, conventional non-convex optimization methods are not proper to be employed to solve these difficulties.


\section{Dueling Double Deep Q-network Algorithm for Trajectories Planning and Passive Beamforming Design}
In this section, we propose a machine learning (ML)-based algorithm, namely, dueling double deep Q-network (D$^{3}$QN)-based algorithm to solve problem \eqref{8}, which is invoked for trajectories planning and the phase shifts of the IRS, as well as the power allocation from the AP to the robots, where the D$^{3}$QN makes full use of the advantages of double deep Q-network (Double DQN) \cite{IEEEhowto:Hasselt} and dueling deep Q-network (Dueling DQN) \cite{IEEEhowto:Wang}.

\subsubsection{State in the $D^{3}$QN Model}
The state space \emph{$\pmb{E} = \{e_{t}\}$} at each epoch of the IRS-enhanced multi-robot networks is defined into three parts: the current phase shift \emph{$\theta_{k}(t) \in [0,2\pi)$} of each passive reflecting elements in the IRS, the current position (\emph{$x_{i}$},\emph{$y_{i}$},\emph{$h_{r}$}) of the robot \emph{$i$}, and the current set of allocation power \emph{$\{p_{i}\}$} from the AP to all the robots. Thus, the state space \emph{$\pmb{E}$} can be expressed as
\begin{align}\label{15}
  \pmb{E} = [\theta_{1}(t),\theta_{2}(t),\cdots,\theta_{K}(t);x_{i};\{p_{i}\}].
\end{align}
\par
The total number of positions \emph{$N_{p}$} in the trajectories is denoted as (\emph{$|x_{\hat{S}_{I}} - x_{\hat{S}_{F}}|+|y_{\hat{S}_{I}} - y_{\hat{S}_{F}}|-1$}) with randomly generated \cite{IEEEhowto:Eckhardt} initial position \emph{$\hat{S}_{I}$} and final position \emph{$\hat{S}_{F}$}. The primary state space complexity is calculated as ($K+N_{p}+3$).

\subsubsection{Action in the $D^{3}$QN Model}
The action space \emph{$\pmb{F} = \{f_{t}\}$} at each epoch of the IRS-enhanced multi-robot networks is defined into three parts: the available quantity of phase shifts \emph{$\{\frac{2\pi n_{0}}{2^{B_{0}}}, n_{0} = 0,1,2,\cdots,2^{B_{0}}-1\}$}, the moving direction and distance \emph{$\pmb{D} = \{d_{r},d_{l},d_{0},d_{u},d_{d}\}$} for the IRS, the available quantity of power allocation \emph{$\{p_{1},p_{2},\cdots,p_{v}\}$}. Note that the \emph{$B_{0}$}, \emph{$d_{g}, \pmb{u}=\{r,l,0,u,d\}$}, \emph{$v$} denote the resolution for the IRS phase shift, the right-left-stillness-up-down direction with 1 unit pace, and the total number of the available power allocated to the robots. Thus, the action state \emph{$\pmb{F}$} can be expressed as
\begin{align}\label{16}
  \pmb{F} = [\{\frac{2\pi n_{0}}{2^{B_{0}}}\};\pmb{D};\{p_{1},p_{2},\cdots,p_{v}\}].
\end{align}
\par
Accordingly, the primary action space complexity is calculated as ($2^{B_{0}}+v+3$).

\subsubsection{Reward in the $D^{3}$QN Model}
The reward is a considerable factor in the optimization of trajectories planning and passive beamforming design. To make the calculation simple, we compare the communication rates between two adjacent timeslots of all the robots and give the robots the 1000 times rate difference as rewards. The reward function of robot \emph{$i$} can be calculated as
\begin{align}\label{17}
  &i(t) = 1000\{[{\rm log}_{2}(1+\frac{|(\pmb{\upsilon})^{H}\pmb{\psi}_{i}+\overline{\pmb{h}}_{i}|^{2}p_{i}}{\sum\limits_{O(j) > O(i)}|(\pmb{\upsilon})^{H}\pmb{\psi}_{j}+\overline{\pmb{h}}_{j}|^{2}p_{j}+\sigma^{2}})]_{(t)} \nonumber \\
  &- [{\rm log}_{2}(1+\frac{|(\pmb{\upsilon})^{H}\pmb{\psi}_{i}+\overline{\pmb{h}}_{i}|^{2}p_{i}}{\sum\limits_{O(j) > O(i)}|(\pmb{\upsilon})^{H}\pmb{\psi}_{j}+\overline{\pmb{h}}_{j}|^{2}p_{j}+\sigma^{2}})]_{(t-1)}\},
\end{align}
\par
\noindent
where $[\cdot]_{(t)}$ and $[\cdot]_{(t-1)}$ denotes the uniform for the communication rates at time \emph{$t$} and \emph{$t-1$}. Thus, the maximization of long-term sum reward can make dedication to the optimize trajectories and passive beamforming. 

\begin{algorithm}[htbp]
  \caption{${\rm D}^{3}$QN-based algorithm for trajectories planning and beamforming design}
  \label{DDDQN}
  \begin{algorithmic}[1]
  \REQUIRE ~~\\
  Q-network structure, LSTM network structure, ARIMA structure, reply memory $\pmb{\mathcal{D}}$, minibatch size $\overline{N}$.\\
  \ENSURE Target Q-value function and decision policy.\\
  \STATE \textbf{Initialize:} 
  \STATE Reply memory $\pmb{\mathcal{D}}$, Q-table $\pmb{Q}$, R-table $\pmb{r}$, state space $\pmb{E}$, action space $\pmb{F}$, 2D space $\pmb{M}$, Q-network weights $\psi$, $\overline{\psi}$, $\overline{\psi}_{1}$, and $\overline{\psi}_{2}$, phase shift $\pmb{\Theta}$, power allocation $\{p_{i}\}$.\\
  \STATE Explore the positions of AP, IRS, and boundaries in $\pmb{M}$.
  \STATE LSTM-ARIMA for initial-final positions prediction.
  \STATE Randomly choose phase shift and power allocation factor.
  \REPEAT
  \STATE The agent randomly selects $e_{t} \in \pmb{E}$ with probability $\epsilon$.\\
  \STATE Execute action $f_{t}$, observe reward $r(t)$, append to $e_{t+1}$.\\
  \STATE Otherwise select ${\rm arg} \max\limits_{f_{t+1}} Q(e_{t+1},f_{t};\psi_{t})$.\\
  \STATE According to the NOMA method, determine the decoding \\
         order for the robots.\\
  \STATE Store transition ($e_{t}$, $f_{t}$; $r(t)$; $e_{t+1}$) in $\pmb{\mathcal{D}}$.\\
  \STATE Sample random minibatch of transition ($e_{t}$, $f_{t}$; $r(t)$; $e_{t+1}$) from $\pmb{\mathcal{D}}$. Set $W_{t} = $\\
  \STATE $\left\{ \begin{array}{lr} r(t), \hspace{1em} {\rm if\ episode\ terminates\ at\ step\ }t+1, & \\ r(t) + \eta Q(e_{t+1},f_{t}^{max}(e_{t+1},\psi_{t});\overline{\psi}_{t},\overline{\psi}_{t_{1}},\overline{\psi}_{t_{2}}),{\rm others} \end{array}\right.$\\ \hspace*{2.5em} 
  \FOR{$u \in \{\psi, \overline{\psi}, \overline{\psi}_{1}, \overline{\psi}_{2}\}$}
      \STATE Perform a gradient descent step:\\
      \STATE $u_{t+1} = u_{t} + \eta_{0}[r(t)+\eta\max_{f_{t+1}}Q(e_{t+1},f_{t+1};u_{t})-Q(e_{t},f_{t};u_{t})]\triangledown_{u_{t}}Q(e_{t},f_{t};u_{t})$.
  \ENDFOR
  \STATE Every $\mathcal{C}$ steps reset $Q_{f}(\cdot)$.\\
  \UNTIL{$\pmb{E}$ is terminal.}
  \end{algorithmic}
\end{algorithm}

\subsubsection{$D^{3}$QN-based Algorithm for trajectories planning and IRS design}
In the $D^{3}$QN model, the AP acts as an agent. The controller is installed at the AP, while the AP can make a decision policy for the robots' positions, IRS phase shifts adjustment, as well as the power allocation from the AP to the robots. At each timeslot \emph{t}, the AP observes the state \emph{$e_{t} \in \pmb{E}$} of the system. The decision policy in the $D^{3}$QN model is determined by Q value in the Q-function. According to the double DQN model, it does not directly find the maximum Q value in each action in the target Q network, but first finds the action corresponding to the maximum Q value in the current Q network, which can be expressed as
\begin{align}\label{18}
  f_{t}^{max}(e_{t+1},\psi_{t})={\rm arg} \max_{f_{t+1}} Q(e_{t+1},f_{t};\psi_{t}),
\end{align}
\par
\noindent
and the target Q-value can be calculated by
\begin{align}\label{19}
  W_{t} = r(t) + \eta Q(e_{t+1},f_{t}^{max}(e_{t+1},\psi_{t});\overline{\psi}_{t}),
\end{align}
\par
\noindent
where the \emph{$\psi_{t}$} and \emph{$\overline{\psi}_{t}$} are the parameters for the action value function and state value function, respectively. The update method for \emph{$\psi_{t}$} and \emph{$\overline{\psi}_{t}$} are identical. For \emph{$\psi_{t}$}, the update equation can be expressed as
\begin{align}\label{20}
  \psi_{t+1} = \psi_{t} + \eta_{0}[r(t)&+\eta\max_{f_{t+1}}Q(e_{t+1},f_{t+1};\psi_{t}) \nonumber \\
  &-Q(e_{t},f_{t};\psi_{t})]\triangledown_{\psi_{t}}Q(e_{t},f_{t};\psi_{t}),
\end{align}
\par
\noindent
where the \emph{$\triangledown_{\{\cdot\}}$} is the gradient operator. However, in some given states, any action has little effect on the state. In order to consider this case, we employ the structure of dueling DQN to measure the value of the state and the value of the action in the state. Thus, the Q-value in the equation \eqref{19} can be rewritten as
\begin{align}\label{21}
  &Q(e_{t+1},f_{t}^{max}(e_{t+1},\psi_{t});\overline{\psi}_{t},\overline{\psi}_{t_{1}},\overline{\psi}_{t_{2}}) \nonumber \\
  &= Q_{e}(e_{t+1};\overline{\psi}_{t},\overline{\psi}_{t_{1}}) + Q_{f}(e_{t+1},f_{t}^{max}(e_{t+1},\psi_{t});\overline{\psi}_{t},\overline{\psi}_{t_{2}}),
\end{align}
\par
\noindent
where the \emph{$Q_{e}(\cdot)$}, \emph{$Q_{f}(\cdot)$}, \emph{$\overline{\psi}_{t_{1}}$}, \emph{$\overline{\psi}_{t_{2}}$} denote the state value function, action value function, parameter for state value function, and parameter for action value function, respectively. Additionally, in order to enhance the identifiability of \emph{$Q_{e}(\cdot)$} and \emph{$Q_{f}(\cdot)$}, we introduce a mean square error (MSE) loss function, which can be given by
\begin{align}\label{22}
  &Q(e_{t+1},f_{t}^{max}(e_{t+1},\psi_{t});\overline{\psi}_{t},\overline{\psi}_{t_{1}},\overline{\psi}_{t_{2}}) \nonumber \\
  &= Q_{e}(e_{t+1};\overline{\psi}_{t},\overline{\psi}_{t_{1}}) + (Q_{f}(e_{t+1},f_{t}^{max}(e_{t+1},\psi_{t});\overline{\psi}_{t},\overline{\psi}_{t_{2}}) \nonumber \\
  &-\frac{1}{\pmb{\mathcal{A}}}\sum_{(f_{t}^{max})^{'}\in \pmb{\mathcal{A}}}Q_{f}(e_{t+1},(f_{t}^{max})^{'}(e_{t+1},\psi_{t});\overline{\psi}_{t},\overline{\psi}_{t_{2}}),
\end{align}
\par
\noindent
where the \emph{$\pmb{\mathcal{A}}$}, \emph{$(f_{t}^{max})^{'}(\cdot)$} are the sampling set and a action sampled in the set, respectively. The target Q-value can be re-calculated as
\begin{align}\label{23}
  W_{t} = r(t) + \eta Q(e_{t+1},f_{t}^{max}(e_{t+1},\psi_{t});\overline{\psi}_{t},\overline{\psi}_{t_{1}},\overline{\psi}_{t_{2}}).
\end{align}
\par
Thus, according to the $D^{3}$QN model mentioned above, through continuous learning, the agent can find the optimal execution policy for the robot trajectories and phases in the IRS. The detailed pseudo code is shown in \textbf{Algorithm~\ref{DDDQN}}.

\begin{remark}\label{remark 1}
  One of the core decisions of the Q-network-based reinforcement learning algorithm is the $\epsilon$-greedy algorithm. When training the model, the robot will have a probability of ($1-\epsilon$) not to perform the expected decision. Therefore, the episodes of convergences to the model are different in each training.
\end{remark}

\section{Numerical Results}
In this section, we provide simulation results to verify the effectiveness of the proposed machine learning-based optimization algorithms for joint trajectories planning and passive beamforming design, as well as the performance of the algorithms. In the simulation, the standard size of length, width, and height in this indoor environment is $8m \times 6m \times 3m$, while there are four pillars with regular size $1m \times 1m \times 3m$ and two parterres with regular size $1m \times 1m \times 1m$. Additionally, in the middle of the 2D plane, an artificial fountain with a regular base size $1.5m \times 1.5m \times 1m$. Note that, all values of the height mentioned above are more than that of the robot. The maximal transmit power at AP is pre-defined as 20 dBm, while the heights of AP and IRS are defined as $2m$ and $3m$. We analyze the performance of the proposed $D^{3}$QN algorithms, the trajectories for all the robots, and the achievable sum-rate for all the robots.

\subsection{The analysis for $D^{3}$QN algorithms}
\begin{figure}[htbp] 
  \centering
  \includegraphics[height=2.4in,width=3.2in]{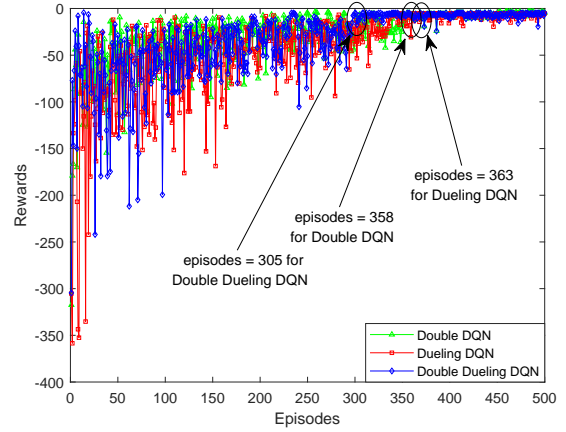}
  \caption{Performance for D$^{3}$QN algorithm comparing different algorithms.}
  \label{Performance for DDDQN algorithm comparing different algorithms}
\end{figure}

The performance of the ML-based algorithm occupies a pivotal place in the entire optimization. In the $D^{3}$QN algorithm proposed in this paper, after selecting the initial-final points and the total path length of the robot, we optimize the sum-rate. In order to analyze the performance of $D^{3}$QN, we compared the double DQN algorithm and the dueling DQN algorithm. As shown in Fig.~\ref{Performance for DDDQN algorithm comparing different algorithms}, the double DQN algorithm and dueling DQN algorithm have no big difference in convergence speed,. They can converge when episodes are 358 and 363 respectively. However, the convergence speed of the proposed $D^{3}$QN is faster than these two algorithms, reaching 305. It is worth noting that in virtue of $\epsilon$-greedy strategy, the convergence episodes of these three algorithms cannot be guaranteed to be the same during each training. Therefore, the result given in the figure is the average convergence given by 10 repetitive training.
\begin{figure}[htbp] 
  \centering
  \includegraphics[height=2.4in,width=3.2in]{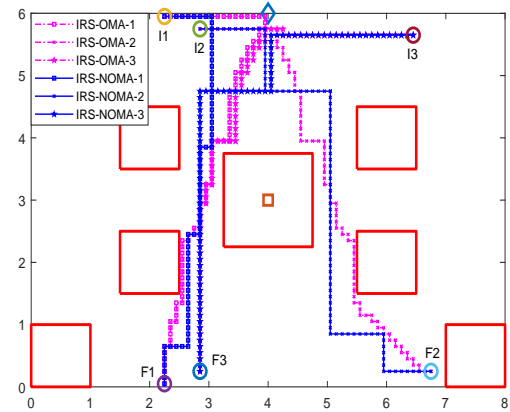}
  \caption{Trajectories for each robot under IRS-OMA and IRS-NOMA cases, elements = 30.}
  \label{path}
\end{figure}
\begin{figure}[htbp] 
  \centering
  \includegraphics[height=2.4in,width=3.2in]{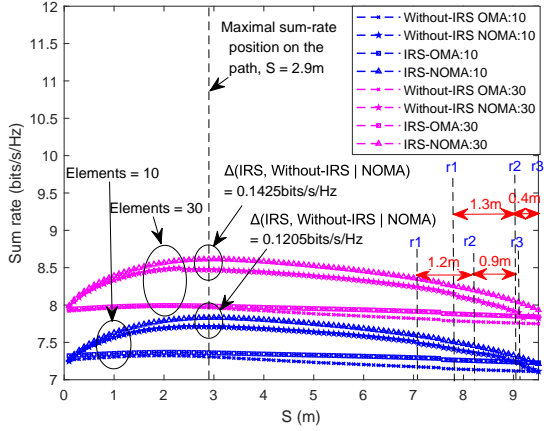}
  \caption{sum-rate versus path length with different elements numbers of IRS.}
  \label{sum-rate versus path length with different elements numbers of IRS}
\end{figure}

\subsection{Achievable sum-rate for the robots}
Denote the velocity of robots and the resolution as 0.1 m/s and 0.1 m, respectively, which can guarantee the path length traversed by each robot is identical at each time slot. Note that the robot can only move back, forth, left, and right. For ease of exposition, we make the size of the grid approximate to the center point of the grid as the resolution is small. Thus, the sum-rate at each timeslot \emph{$t_{s}$} can be calculated when the robots move to the center of the \emph{s}-th grid on their trajectories. As shown in Fig.~\ref{path}, the paths for all robots are depicted in "IRS-OMA" and "IRS-NOMA" cases, while the number of elements in the IRS is 10. The "$\circ$" with "$I_{w},w=\{1,2,3\}$" denotes the initial position for the robots, while the "$\circ$" with "$F_{w},w=\{1,2,3\}$" represents the final position. It is observed that all planned paths tend to be close to the positions of AP and IRS.
\par
As shown in Fig.~\ref{sum-rate versus path length with different elements numbers of IRS}, mark the "OMA-strategy" as a benchmark scheme, the maximal sum-rate for three robots at any given point on their trajectories are obtained. Comparing ”IRS-NOMA” and ”Without-IRS NOMA” cases, the maximal sum-rate difference between ”with IRS deployment” and ”without IRS deployment” cases reaches 0.1425 bits/s/Hz when the IRS elements are 30, while the gap reaches 0.1205 bits/s/Hz when the IRS elements are 10. Additionally, the discrepancy under "with IRS deployment" and "without IRS deployment" cases between the NOMA and OMA strategies from 0.3512 bits/s/Hz to 0.4866 bits/s/Hz, and 0.5236 bits/s/Hz to 0.6490 bits/s/Hz, respectively. Furthermore, Fig.~\ref{sum-rate versus path length with different elements numbers of IRS} shows the path length of each robot from the randomly generated initial position to the final position. In the "IRS-OMA" case, the total path length of each robot is identically achieved 9.5m, whether the number of elements is 10 or 30. When the NOMA strategy is employed, the path lengths of robots are successively 7.1 m, 8.3 m, and 9.2 m with the case of 10 elements in the IRS, respectively. When the elements are increased to 30 elements, the path length of each robot is 7.8 m, 9.1 m, and 9.5 m, respectively. It is worth noting that the first robot has priority to reach the final point, however, the transmission signal it receives cannot be interrupted until other robots arrive at their planned final positions. Furthermore, in order to compare to the "IRS-OMA" case, the signal is an outage at 9.5s, which equivalent to the total path length of 9.5 m. The sum-rate at each timeslot can be calculated according to the equation \eqref{7}.

\section{Conclusion}
In this paper, we explored a downlink IRS-aided multi-robot NOMA networks. The sum-rate maximization problem was formulated by jointly optimizing trajectories for robots, reflecting coefficients matrix, the decoding order, and the power allocation at the AP, subject to the QoS for all the robots. To tackle the formulated problem, a machine learning algorithm were proposed to plan trajectories for the robots and design the phase shift matrix. Numerical results were provided for demonstrating that the proposed IRS-aided NOMA networks achieve significant gain compared to IRS-OMA and without-IRS assisted scheme. Additionally, the explored D$^{3}$QN algorithm attained considerable performance compared to the conventional algorithm.

\appendices

\ifCLASSOPTIONcaptionsoff
  \newpage
\fi

\end{document}